\documentclass{article}
\usepackage{spconf,amsmath,graphicx,amsfonts, hyperref}
\usepackage{xcolor}
\usepackage{multirow}
\usepackage{booktabs}
\usepackage{makecell}
\usepackage[inline]{enumitem}

\newcommand{\rescell}[2]{\makecell{$#1{\scriptstyle \pm#2}$}}
\newcommand{\namecell}[1]{#1}

\title{VA\MakeLowercase{u}LT: Augmenting the Vision-and-Language Transformer\\for Sentiment Classification on Social Media}

\name{Georgios Chochlakis$^1$\quad Tejas Srinivasan$^2$ \quad Jesse Thomason$^2$ \quad Shrikanth Narayanan$^1$}
\address{\normalsize $^1$ Signal Analysis and Interpretation Lab, University of Southern California, Los Angeles, CA 90089, USA \\ \normalsize $^2$ GLAMOR Lab, University of Southern California, Los Angeles, CA 90089, USA }

\begin{document}
\ninept
\maketitle
\begin{abstract}
We propose the \textbf{Vision-and-Augmented-Language Transformer} (\textbf{VAuLT}).
VAuLT is an extension of the popular Vision-and-Language Transformer (ViLT), and improves performance on vision-and-language (VL) tasks that involve more complex text inputs than image captions while having minimal impact on training and inference efficiency.
ViLT, importantly, enables efficient training and inference in VL tasks, achieved by encoding images using a linear projection of patches instead of an object detector. 
However, it is pretrained on captioning datasets, where the language input is simple, literal, and descriptive, therefore lacking linguistic diversity.
So, when working with multimedia data in the wild, such as multimodal social media data, there is a notable shift from captioning language data, as well as diversity of tasks. We indeed find evidence that the language capacity of ViLT is lacking.
The key insight and novelty of VAuLT is to propagate the output representations of a large language model (LM) like BERT to the language input of ViLT.
We show that joint training of the LM and ViLT can yield relative improvements up to $20\%$ over ViLT and achieve state-of-the-art or comparable performance on VL tasks involving richer language inputs and affective constructs, such as for Target-Oriented Sentiment Classification in TWITTER-2015 and TWITTER-2017, and Sentiment Classification in MVSA-Single and MVSA-Multiple. Our code is available at \url{https://github.com/gchochla/VAuLT}.
\end{abstract}

\vspace{-0.1cm}

\vspace{-0.5cm}
\section{Introduction}
\label{sec:intro}

\begin{figure}[t]
\centering
\includegraphics[width=0.9\columnwidth]{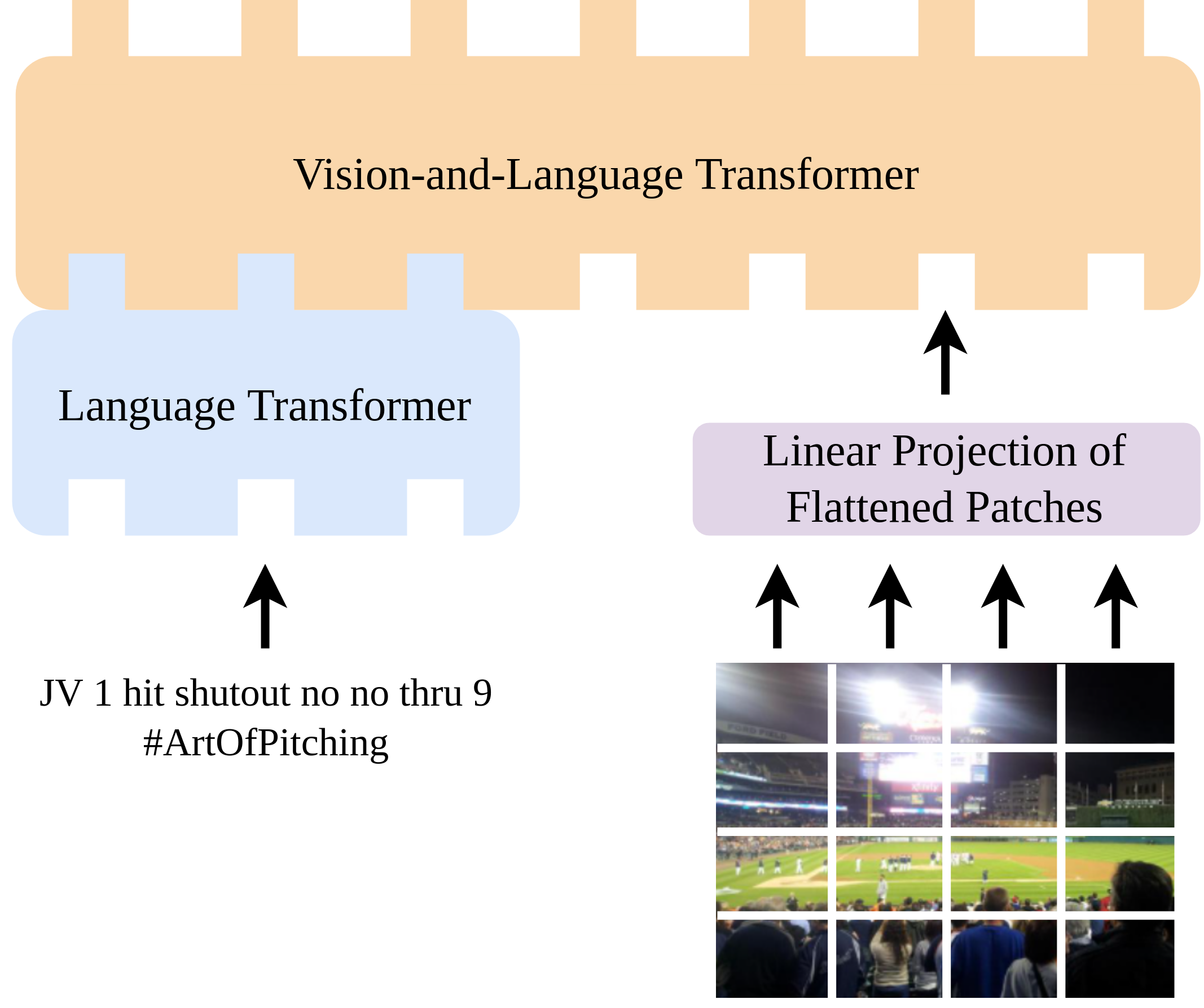}
\caption{VAuLT propagates representations from a LM to a VLM in a modular fashion. We show that even with little tuning data, language representations can be effectively used to significantly improve the performance of ViLT on tasks with greater linguistic and construct diversity than captioning datasets.}
\label{fig:thumbnail}
\end{figure}

The study of social media can have positive societal outcomes. 
Interest in the importance of emotion has been growing beyond science \cite{dukes2021rise} to law \cite{wistrich2014heart}, and politics \cite{wahl2019emotions}. 
Social media users readily express their emotions about their experiences and world events through multimedia content such as videos, images, memes, and raw text.
This allows researchers to use machine learning (ML) methods to study, for instance, the dynamics of public perception and opinion and to provide feedback to policy makers and inform future actions \cite{hoover2018moral}. 
Social media are also ripe with opportunities for social manipulation \cite{bradshaw2018global}. This manipulation can be personalized based on user data \cite{alfano2022ethical}, or use network effects \cite{hoover2021investigating}. Mitigating these risks and identifying widespread manipulation attempts hence becomes pivotal.

Due to the massive volume of data and the dynamic nature of social media, constant streams of data are required, thus efficiency is of the essence. In addition, multimedia content necessitates multimodal modeling, as users can use these extra degrees of freedom to convey complex and conflicting meanings, for instance in irony. Moreover, a multimodal model should be able to handle a subset of its input modalities if a post is missing the rest \cite{srinivasan2022climb}.

Many state-of-the-art vision-and-language models (VLM), e.g., VL-BERT \cite{su2019vl} and ViLBERT \cite{lu2019vilbert}, contain an object detector, such as Faster-RCNN \cite{ren2015faster}, to compute input visual features. Object detectors are costly to run relative to the rest of the pipeline \cite{kim2021vilt}, which can be prohibitive for social media applications. Hence, we turn to the Vision-and-Language Transformer (ViLT)~\cite{kim2021vilt}, whose visual input consists of linearly projected image patches.

Regardless, these models, designed to understand and extract information from multimedia content such as vision and language, are frequently trained \textit{only} on images and their captions crawled from the web.
Image captions are a weak form of language supervision, often describing literal content with simple syntax and little subtlety, deviating substantially from the text of social media posts.

In this paper, we propose the Vision-and-Augmented-Language Transformer (VAuLT), which directly addresses the impoverished language representations of ViLT by processing the language input through a pretrained large language model (LM) (Figure~\ref{fig:thumbnail}).
We find that VAuLT is able to outperform ViLT on a variety of tasks involving affective constructs common in social media, including TWITTER-2015, TWITTER-2017 \cite{yu2019adapting}, and MVSA \cite{MVSA}.

Our key contributions can be summarized as follows:
\begin{itemize}
    \item In VAuLT, we stack Transformer architectures, a pretrained LM and VLM, and jointly train them end-to-end, yielding substantial performance gains from little tuning data.
    \item We show that VAuLT competes with or outperforms state-of-the-art methods for the social media analysis tasks examined.
    \item We provide evidence that the performance trade-offs of ViLT and VAuLT depend on the complexity of the image and language components commensurate with the relative richness of, and variety within, the modalities in the source and target domains, as well as the tasks themselves.
\end{itemize}

\section{Related Work}
\label{sec:related}

\subsection{Language Transformers}
Transformers pretrained on general-purpose text grants them the ability to be used as backbones for many downstream tasks \cite{devlin2018bert}. Such models can also be pretrained on more specialized corpora to improve their transferability to specific domains, like BERTweet \cite{nguyen2020bertweet}, a model trained on tweets.

\subsection{Vision-and-Language (VL) Transformers} 
Similar to language transformers, VL transformers \cite{su2019vl,lu2019vilbert} are trained with multiple input modalities on web-crawled captions, and yield state-of-the-art results on many multimodal tasks \cite{antol2015vqa}.
These models frequently use region-of-interest (ROI) features extracted from a pretrained object detector \cite{ren2015faster} as their Transformer's visual inputs. 
However, this visual embedding step is computationally expensive during training and inference \cite{kim2021vilt}, requiring caching of features. 
ViLT~\cite{kim2021vilt}, based on the Vision Transformer (ViT)~\cite{dosovitskiy2020image}, proposed using image patches as visual input to the Transformer model instead. 
This results in slightly worse model performance, but is much more computationally efficient. 
However, ViLT also has limited language understanding capacity due to being initialized from ViT and pretrained solely on image caption data. 
VAuLT overcomes this issue by replacing ViLT's language embedding inputs with features extracted from a large LM pretrained on greater linguistic diversity, which can also be selected to more closely suit the needs of the downstream task.

\subsection{Affective Analysis on Social Media}

Researchers tend to focus on single modalities when analyzing emotion and sentiment in social media posts, such as text-only \cite{mohammad2018semeval}, and image-only \cite{sharma2020semeval} datasets.

Emotion recognition from text has long relied on word-counting techniques \cite{garten2018dictionaries}, which can be deployed at scale but disregard context. LSTMs \cite{baziotis2018ntua} and Transformers \cite{chochlakis2022leveraging} capture context and provide superior performance with less interpretability.

On the other hand, multimodal affect modeling tasks with social media have predominantly focused on sentiment \cite{yu2019adapting, MVSA, hu2018multimodal}. 
\cite{yu2019adapting} have annotated tweets with their corresponding images for sentiment expressed towards entities in the text, resulting in the \textit{Target-oriented Sentiment Classification} (TMSC) task and the TWITTER-15 and TWITTER-17 datasets. Similarly, MVSA \cite{MVSA} has annotations for sentiment expressed in the text and the image of a tweet. User-provided tags have been used as ``self-reports'' of emotion in scraped multimodal data from Tumblr \cite{hu2018multimodal}.

The proposed approaches for the aforementioned tasks typically involve dedicated architectures or targeted modifications of VL Transformers. An example of the former is TomBERT \cite{yu2019adapting} for TMSC. TomBERT consists of multiple components: first, a ResNet \cite{he2016deep} extracts visual features. Then, a cross-attention layer is used to query these visual features using target embeddings extracted from BERT \cite{devlin2018bert}. Parallel to the visual feature extraction, the text sequence is passed though another BERT, and is concatenated with the refined visual features and fed to a multimodal encoder, also a BERT.

\section{Methodology}
\label{sec:method}

\subsection{Task Definitions}

\subsubsection{Sentiment Classification} For our purposes, ``Sentiment Classification'' corresponds to classifying an image and text pair as \textit{positive}, \textit{neutral}, or \textit{negative}.

\vspace{-0.4cm}
\subsubsection{Target-oriented Sentiment Classification (TMSC)} TMSC extends the basic sentiment classification formulation to opinion expressed toward {\it given} targets within the text of multimodal tweets.
In our setting, sentiment targets were repurposed from types of named entities, and an input tweet can contain multiple targets.

\subsection{Vision-and-Language Transformer (ViLT)} 

ViLT is a multimodal transformer. The language component of ViLT follows conventional practices and embeds input tokens with a look-up table. On the other hand, the visual component of ViLT follows ViT \cite{dosovitskiy2020image}, segmenting each image into non-overlapping square patches, flattening the resulting values, and mapping them to the input space using a linear projection.

ViLT is pretrained on image captioning datasets like MSCOCO \cite{lin2014microsoft}. Its pretraining objectives include: image-text matching and word-patch alignment, and masked language modeling and whole word masking. The former group of objectives focuses on enforcing joint modeling of the two modalities, while the latter is centered on the language understanding of the model. Explicit language training is necessary since the model is initialized from ViT weights, itself trained on images, in contrast to previous works that initializes weights from LMs \cite{su2019vl, lu2019vilbert}, thus \textit{ViLT's language understanding comes solely from image-captioning datasets}.

\subsection{Vision-and-Augmented Language Transformer (VAuLT)}

The proposed VAuLT can use any LM that produces token embeddings for tokenized text, irrespective of the specific tokenizer. We have this requirement because we use the sequence of embeddings produced by the LM to replace the canonical language input of ViLT. Therefore, we are essentially substituting the context-free representations from the lookup table of ViLT, with context-aware representations \cite{devlin2018bert} from the output layer of a LM. Images are embedded in the canonical fashion with ViLT's trained linear projection of patches.

Concretely, given an input image-text pair, and a specific LM and its tokenizer, we tokenize the input text sequence and pass this through the LM to produce its output contextual embeddings. Along with these embeddings, we pass these two input streams through ViLT. Note that position embeddings, etc., are added in both ViLT's and the LM's forward pass.

\begin{table*}
    \centering
    \begin{tabular}{lcccccccc}
        & \multicolumn{2}{c}{TWITTER-15} & \multicolumn{2}{c}{TWITTER-17} & \multicolumn{2}{c}{MVSA-M} & \multicolumn{2}{c}{MVSA-S} \\
        \cmidrule(lr){2-3} \cmidrule(lr){4-5} \cmidrule(lr){6-7} \cmidrule(lr){8-9}
        Model & Acc & mac-F1 & Acc & mac-F1 & Acc & w-F1 & Acc & w-F1 \\
        \midrule
            \namecell{ViLT} & \rescell{70.5}{1.3} & \rescell{62.6}{2.5} & \rescell{62.6}{0.1} & \rescell{58.1}{0.7} & \rescell{69.1}{0.8} & \rescell{67.0}{0.3} & \rescell{74.4}{1.5} & \rescell{73.7}{1.6} \\
            
            \namecell{VAuLT$^\dagger$} & \rescell{75.6}{0.8} & \rescell{70.0}{1.7} & \rescell{70.2}{0.4} & \rescell{67.8}{0.1} & \rescell{70.0}{0.7} & \rescell{68.9}{0.7} & \rescell{\mathbf{78.0}}{1.2} & \rescell{77.4}{1.4} \\
            
            \namecell{VAuLT$^\star$} & \rescell{\mathbf{77.5}}{0.4} & \rescell{\mathbf{72.9}}{0.5} & \rescell{71.0}{0.5} & \rescell{\mathbf{69.5}}{0.7} & \rescell{70.3}{1.6} & \rescell{67.0}{2.9} & \rescell{72.8}{5.6} & \rescell{71.8}{5.8} \\

            \namecell{EF-CaTr-BERT \cite{khan2021exploiting}} & \rescell{\mathbf{77.9}}{0.8} & \rescell{\mathbf{73.9}}{0.8} & \rescell{\mathbf{72.3}}{0.3} & \rescell{\mathbf{70.2}}{0.2} & - & - & - & - \\

            \namecell{ITIN \cite{zhu2022multimodal}} & - & - & - & - & \textbf{73.5} & - & 75.2 & - \\
    \end{tabular}
    \caption{Direct comparison between ViLT, VAuLT ($^\dagger$: w/~BERT, \texttt{bert-base-uncased}, or $^\star$: w/~BERTweet, \texttt{vinai/bertweet-base}), and the state-of-the-art on each benchmark. \textbf{Bold} indicates best performance based on ranges, where available.}
    \label{tab:main}
\end{table*}

We use LMs like BERT and BERTweet that have not been trained on similar datasets with ViLT or tokenize text in the same manner as ViLT does. That is to say, the language tokens might not correspond to the tokens ViLT has been trained with.

\section{Experiments} \label{sec:exp}

Overall, our goal is to show that VAuLT improves upon ViLT on multimodal, small-scale, affective datasets derived from social media, in our case Twitter. More than that, we also show that ViLT can achieve competitive performance over dedicated architectures for such datasets, and VAuLT can even exceed their performance.

\subsection{Datasets}

We use TWITTER-15 and TWITTER-17 \cite{yu2019adapting} annotated for the task of TMSC. The former contains 1548 positive, 630 negative and 3169 neutral examples. The latter contains 2516 positive, 728 negative and 2728 neutral examples. We use the splits provided by the authors. Accuracy and macro F1 are typical evaluation criteria. We predict for one target at a time, even if the tweet contains multiple. We follow \cite{yu2019adapting} in replacing the target of interest with the placeholder \$T\$, and appending the target as a second sequence.

We also evaluate our model on MVSA-Multiple and MVSA-Single \cite{MVSA}, pre-processed for Sentiment Classification. Initially, both datasets contain annotations for the sentiment of both the image and the text of a tweet. MVSA-Single has annotations from solely one annotator, while MVSA-Multiple utilizes three separate annotators, plus more examples.
We merge the annotations into a single annotation and aggregate across annotators, as is customary \cite{zhu2022multimodal}. 
The filtered MVSA-Single contains 2683 positive, 460 negative and 1358 neutral examples. The filtered MVSA-Multiple contains 11318 positive, 4408 negative and 1298 neutral examples.
We also had to remove 3 additional samples from MVSA-Multiple because of corrupted images (example IDs: 3151, 3910 and 5995). Since canonical splits are not provided and evaluation criteria are not well-defined, we follow previous work in randomly splitting the data 8:1:1 \cite{zhu2022multimodal} and use accuracy to compare. We also include the weighted F1 score of our methods as our best guess as to what other methods are reporting as merely ``F1 score''.

\subsection{Implementation details}

We use Python v3.7.4, PyTorch v1.11.0, and \textit{transformers}  v4.19.2. 

We keep our learning rate at $2\cdot10^{-5}$ unless specified. We use linear warm-up for $10\%$ of the training steps and then linear decay to~$0$. In re-implementations of published models, we use the specified hyperparameter configuration.

Test performance (mean and standard deviation) is reported after 3 different training runs with fixed hyperparameters. The main tuning of all of the hyperparameters but the number of training epochs was performed on TWITTER-15 and TWITTER-17. We searched for whether to use bias correction in AdamW, selecting no bias correction, the number of epochs (ran for 15 epochs and picked the best performing one based on average accuracy and macro F1 for MVSA-Multiple and MVSA-Single, and $\{8, 15, 25\}$ for TWITTER-15 and TWITTER-17), and whether to integrate the placeholder \$T\$ to the tokenizer as a standalone token in TWITTER-15 and TWITTER-17 (yes for VAuLT; no for ViLT, existing baselines, and variants thereof).

We find that divergence in the training of VAuLT can rarely occur, but those runs can be filtered by observing degradation in the training metrics. While we had to correct for such artifacts during development, no such instances occurred on the test sets. 

We use simple random cropping instead of the augmentation utilized in ViLT for simplicity. We use augmentation in all cases where ViLT's linear projection is solely used to extract image embeddings. We use \textit{ekphrasis}\footnote{https://github.com/cbaziotis/ekphrasis} to pre-process the text of the tweets, but we remove the special characters added by the library around the names of the tags (e.g., when ``$<$user$>$'' would have been the original substitute for the actual Twitter handle appearing in a tweet, we simple use ``user''). We also substitute emojis for their plain-text descriptions in parentheses, when they would be dropped by a tokenizer. In particular, in our work, only BERTweet's tokenization preserves emojis. Because the text of TWITTER-15 and TWITTER-17 has been independently pre-processed, we do not use \textit{ekphrasis} on them.

We use 2 NVIDIA GeForce GTX 1080 Ti GPUs (12GB of memory). We only use one GPU per model, limiting our training batch sizes to 16 or 32, always choosing the maximum possible of the two.

\subsection{Comparison with ViLT}

First and foremost, let us focus on the comparison of the evaluation metrics between ViLT and VAuLT, where we indeed find that VAuLT exceeds ViLT's performance across all examined affective social media data. Results can be seen in Table \ref{tab:main}. We can see that VAuLT with BERTweet improves upon ViLT in TWITTER-15 and TWITTER-17, with relative improvements between $9.9\%$ and $19.6\%$. VAuLT with BERT also provides substantial improvements but still lags behind compared to BERTweet, and only we find overlapping ranges across the two versions of VAuLT on TWITTER-17's accuracy.

Smaller yet still substantial improvements are observed in MVSA-Multiple and MVSA-Single. In particular, for this set of benchmarks, VAuLT with BERT performs favorably compared to ViLT and VAuLT with BERTweet. In fact, for the latter, we observe large fluctuations in performance, emphasized by the large deviation presented in Table \ref{tab:main}. Otherwise, performance is analogous to VAuLT's with BERT. Quantitatively, improvements over ViLT are up to approximately 5\%. It has to be noted that ranges are overlapping for VAuLT and ViLT in MVSA-Multiple accuracy.


\subsection{Comparisons with State of the Art}

For completeness, we also compare VAuLT and ViLT with state-of-the-art models for the examined benchmarks. We do not elaborate on the state-of-the-art models because of the variety of tasks, as well as them being not central to the main goals of our work. We find that VAuLT can outperform or be competitive with dedicated state-of-the-art models in terms of our evaluation criteria.

We observe, in Table \ref{tab:main}, that VauLT with BERTweet has overlapping performance ranges with the state-of-the-art model, EF-CaTr-BERT, on TWITTER-15 and TWITTER-17, except for accuracy in TWITTER-17.

For MVSA-Multiple and MVSA-Single, we have mixed results. We achieve state-of-the-art results for MVSA-Single, where we improve upon the previous best accuracy by 3.8 absolute percentage points (note that our best-guess F1 score is also the state of the art if our assumption is correct). In MVSA-Multiple, our models are still outperformed by the competition. Caveats in the evaluation in MVSA include the random splits that are used in the literature, often with different cardinalities in the different splits, while evaluation criteria remain opaque, and pre-processing, aggregation and normalization techniques could differ between different works.

\subsection{Ablation Studies}

\subsubsection{Frozen LM}

\begin{table}
    \centering
    \begin{tabular}{lcccc}
        & \multicolumn{2}{c}{TWITTER-15} & \multicolumn{2}{c}{TWITTER-17} \\
        \cmidrule(lr){2-3} \cmidrule(lr){4-5}
        Model & Acc & mac-F1 & Acc & mac-F1 \\
        \midrule
            \namecell{ViLT} & \rescell{70.5}{1.3} & \rescell{62.6}{2.5} & \rescell{62.6}{0.1} & \rescell{58.1}{0.7} \\
            \namecell{VAuLT$^\dagger$} & \rescell{75.6}{0.8} & \rescell{70.0}{1.7} & \rescell{\mathbf{70.2}}{0.4} & \rescell{67.8}{0.1} \\
            \namecell{VAuLT$^{\star, f}$} & \rescell{68.8}{0.6} & \rescell{58.8}{1.5} & \rescell{64.0}{0.9} & \rescell{61.0}{0.6} \\
            \namecell{VAuLT$^\star$} & \rescell{\mathbf{77.5}}{0.4} & \rescell{\mathbf{72.9}}{0.5} & \rescell{\mathbf{71.0}}{0.5} & \rescell{\mathbf{69.5}}{0.7} \\
    \end{tabular}
    \caption{Comparing ViLT and VAuLT ($^f$: using a frozen LM, and $^\dagger$: w/~BERT, \texttt{bert-base-uncased}, or $^\star$: w/~BERTweet, \texttt{vinai/bertweet-base}) on TWITTER-15 and TWITTER-17. \textbf{Bold} indicates best based on ranges.}
    \label{tab:frozen_twitter1x}
\end{table}

We first examine whether ViLT can ``read'' the outputs of a LM without any adjustments on the part of the latter. That is to say, we keep the LMs frozen. \cite{merchant2020happens} showed that early layers of transformers change significantly less than later layers during fine-tuning, implying that fine-tuning of the LM in VAuLT may be unnecessary. However, we find joint training to be essential.

Results can be seen in Table \ref{tab:frozen_twitter1x}. Overall, fine-tuning of the language model appears essential, since we see such notable degradation in performance, as VAuLT with a frozen LM drops to the level or even below ViLT. This demonstrates the necessity of fine-tuning the LMs in conjunction with ViLT.

\subsubsection{Deep Vision vs. Deep Language}
In this section, we present an experiment supporting the hypothesis that the complexity of the modality-specific components and their effect on the final performance depends on the domain addressed, and that the linear projections used by ViLT and ViT do not inherently restrict the model's performance. We do so by introducing and studying two variants of TomBERT we call TomVAuLT and TomViLT, solely for the purposes of examining our hypothesis in this section.

TomVAuLT extends TomBERT by replacing the multimodal encoder with ViLT. The multimodal encoder in TomBERT is responsible for the final classification, and its input consists of the queried features from ResNet and the contextual embeddings of the tweet from BERT. In this way, we replace the visual embeddings of ViLT with ResNet features, while we also retain the BERT that pre-processes the tweet before the multimodal encoder, giving us VAuLT in that regard. By disposing of that BERT from TomVAuLT, we arrive at the other variant, TomViLT, where only ViLT is used to process the text inputs. We compare these with ViLT and VAuLT using BERT on the dev sets of TWITTER-15 and TWITTER-17. In this manner, we present all possible configurations:
\begin{enumerate*}[label={\roman*)}]
  \item ViLT: no deep encoder,
  \item VAuLT: deep language encoder,
  \item TomViLT: deep visual encoder,
  \item TomVAuLT: both deep encoders.
\end{enumerate*}

Our results, shown in Table \ref{tab:toms}, demonstrate that having a deep visual encoder actually hurts performance in the presence of a deep language encoder, contrasting previous assumptions that transferring a ``strong'' visual vocabulary is essential, which to the best of our knowledge is a novel analytical result.

\begin{table}
    \centering
    \begin{tabular}{lcccc}
        & \multicolumn{2}{c}{TWITTER-15} & \multicolumn{2}{c}{TWITTER-17} \\
        \cmidrule(lr){2-3} \cmidrule(lr){4-5}
        Model & Acc & mac-F1 & Acc & mac-F1 \\
        \midrule
            \namecell{TomViLT} & \rescell{73.2}{0.7} & \rescell{66.1}{1.3} & \rescell{67.3}{0.5} & \rescell{64.4}{0.7}  \\
            \namecell{ViLT} & \rescell{69.6}{1.0}& \rescell{60.3}{1.5} & \rescell{64.0}{1.2} & \rescell{58.0}{1.5} \\
            \namecell{TomVAuLT} & \rescell{73.2}{0.5} & \rescell{67.1}{0.4} & \rescell{67.3}{0.9} & \rescell{61.1}{0.9} \\
            \namecell{VAuLT} & \rescell{\mathbf{75.0}}{0.6} & \rescell{\mathbf{69.0}}{0.9} & \rescell{\mathbf{67.8}}{1.0} & \rescell{\mathbf{64.9}}{0.9} \\
    \end{tabular}
    \caption{Comparing ViLT, VAuLT, TomViLT and TomVAuLT on TWITTER-15 and TWITTER-17. \textbf{Bold} indicates best based on means only.}
    \label{tab:toms}
\end{table}

\section{Conclusion}
In this work, we introduce VAuLT. It utilizes a large pretrained LM, such as BERT, to propagate enhanced language representations to ViLT in order to perform multimodal tasks. We find that this pre-processing to be integral, as ViLT's impoverished language representations, owning to its limited language exposure during pretraining, cannot be easily fine-tuned to perform out-of-distribution tasks, such as affective analysis on multimodal social media data. Our approach obviates the requirements for extensive pretraining of ViLT on the desired domains, as the appropriate LMs can bridge the distribution shift between source and target. Importantly, we demonstrate state-of-the-art or competitive performance of VAuLT.

Our work opens up several research directions. Anecdotal evidence from experiments not presented in this work indicate VAuLT struggles with pure reasoning tasks compared to ViLT, a direction left for future research. Second, we observe rare divergences during training. This hints at possible misalignment of the output space of the LM and the input space of ViLT. Third, we offer some evidence that shallow visual encoders but deep language encoders can achieve better classification performance to alternatives, a paradigm that could be investigated in greater detail. Finally, VAuLT can be used in multilingual settings with the proper LM choice.

Despite the utility of multimodal emotion recognition on social media, as discussed earlier, multiple ethical concerns can be raised. These include utilization of these models for the identification and suppression of dissent and free speech beyond text, such as in the form of images and memes, as well as their general integration in the aforementioned social manipulation tools, rendering them more powerful with the inclusion of visual capabilities.

\section{Acknowledgements}

We thank Kristina Lerman and Keith Burghardt for their valuable guidance. This project was funded in part by DARPA under contract HR001121C0168.

\vfill\pagebreak
\bibliographystyle{acm}
\bibliography{refs}

\end{document}